\pdfoutput=1
\documentclass[11pt]{article}

\usepackage{EACL2023}

\usepackage{times}
\usepackage{latexsym}
\usepackage{graphicx}
\usepackage{microtype}
\usepackage{multirow}
\usepackage{tikz}
\usepackage{array}
\usepackage{calc}
\usepackage{inconsolata}
\usepackage{pgfplots}
\pgfplotsset{compat=1.17}
\usepackage[T1]{fontenc}
\usepackage[utf8]{inputenc}

\usepackage{microtype}

\usepackage{inconsolata}

\newcommand\blfootnote[1]{%
  \begingroup
  \renewcommand\thefootnote{}\footnote{#1}%
  \addtocounter{footnote}{-1}%
  \endgroup
}

\title{MasonPerplexity at Multimodal Hate Speech Event Detection 2024: Hate Speech and Target Detection Using Transformer Ensembles}

 \author{Amrita Ganguly\textsuperscript{*}, Al Nahian Bin Emran\textsuperscript{*}, Sadiya Sayara Chowdhury Puspo \\ {\bf  Md Nishat Raihan, Dhiman Goswami, Marcos Zampieri} \\ George Mason University, USA \\
 \texttt{\{agangul, abinemra\}@gmu.edu}}

\begin{document}
\maketitle
\begin{abstract}
The automatic identification of offensive language such as hate speech is important to keep discussions civil in online communities. Identifying hate speech in multimodal content is a particularly challenging task because offensiveness can be manifested in either words or images or a juxtaposition of the two. This paper presents the \textit{MasonPerplexity} submission for the Shared Task on Multimodal Hate Speech Event Detection at CASE 2024 at EACL 2024. The task is divided into two sub-tasks: sub-task A focuses on the identification of hate speech and sub-task B focuses on the identification of targets in text-embedded images during political events. We use an XLM-roBERTa-large model for sub-task A and an ensemble approach combining XLM-roBERTa-base, BERTweet-large, and BERT-base for sub-task B. Our approach obtained 0.8347 F1-score in sub-task A and 0.6741 F1-score in sub-task B ranking $3^{rd}$ on both sub-tasks. \blfootnote{\bf * denotes equal contribution.} \blfootnote{\textcolor{red}{\bf This paper contains offensive examples.}}
%The paper underscores the significance of multimodal hate speech detection, showcasing the models' effectiveness in automated identification, while acknowledging ongoing challenges in this evolving research domain.
\end{abstract}

\section{Introduction}

In the context of polarized political discussions, when feelings and perspectives are strong, identifying offensive content is essential to moderation efforts in online communities. The challenge is increased by the use of text-embedded images in which negative emotions can be expressed both verbally and visually. Besides, in the current era of vlogging and reels, people are inclined to utilize memes and emojis or opt for text-embedded images to express their sentiments and comment on online content. As a result, the task of detecting hate speech is expanding to encompass images, posing a new challenge beyond the realm of textual content and across diverse languages.

The Shared Task on Multimodal Hate Event Detection at CASE 2024 \cite{thapa2024multimodal} deals with the identification of hate speech and its targets in text-embedded images during political events. The main objective is to automatically determine if an image that includes text contains hate speech (sub-task A) and, if so, to identify its targets categorized as community, individual, and organization (sub-task B). Identifying the target of offensive messages is vital to understanding their potential harm as demonstrated by annotation taxonomies such as OLID \cite{zampieri2019predicting} and TBO \cite{zampieri2023target}. 

In this paper, we discuss transformer-based approaches to hate speech detection in political events using the Multimodal Hate Speech Event Detection dataset \cite{bhandari2023crisishatemm}. The paper sheds light on the challenges of handling multimodal content, particularly text-embedded images. For sub-task A (hate speech detection), we employ the XLM-roBERTa-large \cite{xlmr} model. For sub-task B (target detection), we adopt an ensemble approach combining XLM-roBERTa-base, BERTweet-large  \cite{berttweetner1}, and BERT-base \cite{bertbaseuncase}. These models are selected to effectively address the unique challenges posed by diverse multimodal content. We report that our approach obtained a 0.8347 F1-score in sub-task A and a 0.6741 F1-score in sub-task B, ranking $3^{rd}$ on both sub-tasks. 

\section{Related Work}

\paragraph{Offensive Content and Hate Speech} Offensive content is pervasive in social media motivating the development of systems capable of recognizing it automatically. While definitions may vary, hate speech is arguably the most widely explored type of offensive content \cite{schmidt2017survey,fortuna2018survey}. Several studies have proposed new datasets and models to label hateful posts on social media \cite{davidson2017,zia2022improving}. More recently, studies have focused on recognizing the specific parts of an instance that may be considered offensive or hateful, as in the case of HateXplain \cite{mathew2020hatexplain},  TSD \cite{pavlopoulos-semeval}, and MUDES \cite{ranasinghemudes}. The vast majority of work on text-based hate speech detection is on English but several papers have created resources and models for languages such as Bengali \cite{raihan2023offensive}, French \cite{chiril-etal-2019-multilingual}, Greek \cite{pitenis-etal-2020-offensive}, Marathi \cite{gaikwad-marathi}, and Turkish \cite{coltekin-2020}.

\paragraph{Multimodal Hate Speech} While the aforementioned studies have focused on the identification of hateful content in texts, there has been growing interest in identifying hateful content in text and images simultaneously. \citet{hermida2023detecting}, \citet{ji2023identifying}, and \citet{yang2022multimodal} highlight the significance of multimodal analysis offering a comprehensive overview of various methodologies employed to detect hate speech in images and memes. Various datasets have been introduced for multimodal hate speech detection \cite{grimminger2021hate,bhandari2023crisishatemm,thapa-etal-2022-multi}  The study by \citet{grimminger2021hate} presents a Twitter corpus with content related to the US elections of 2020. The study by \citet{Boishakhi_2021} explores the combination of various modalities for hate speech detection such as text, video, and audio. While the clear majority of studies deal with English, research on different languages \cite{karim2022multimodal,rajput2022hate,perifanos2021multimodal}. 

%Together, these studies emphasize the evolving landscape and the necessity for sophisticated algorithms capable of comprehending both textual and visual elements for effective hate speech detection across diverse scenarios.
%\citet{hsdnlp-parihar-2021} emphasizes the implementations and barriers of NLP in the field of hate speech detection. \citet{article} briefly discuss about various models, including CNN, BERT, FCM, SCM, TKM, ELMO, and deep belief networks which are employed in hate speech identification with distinct merits and demerits. CNN is efficient for spatial hierarchies but may struggle with long-range dependencies. BERT captures context effectively but has higher time complexity. Feature combination models (FCM, SCM, TKM) enhance performance but introduce complexity. ELMO captures word relationships but requires more computational resources. 

\paragraph{Related Shared Tasks} \citet{thapa-etal-2023-multimodal} organizes CASE 2023, a series of shared tasks identifying Multimodal Hate Speech Event Detection. There are two sub-tasks to identify hate speech and targets in the different sub-tasks. Participants present the utilization of transformer models like BERT, RoBERTa, and XLNet, as well as effective approaches such as vision transformers and CLIP which contributed to the outstanding outcomes. Similarly, different shared tasks have been organized to identify offensive language from texts i.e. \cite{aragon2019overview}, \cite{modha2021overview}. All of this research highlights how important it is to combine several data modalities in order to improve hate speech or offensive language detection.
%

% ST1
% train - nohate: 1942, hate:1658
% tot: 3600
% no hate= 53.94
% hate= 46.05

%  % individual, community, and organization labels are given as 0, 1, and 2 respectively.
% ST2 train - individual: 823, community:335, original:784
% tot: 1942
% ind= 42.37 & 41.80
% com= 17.25 & 16.39
% ori= 40.37 & 41.80

\section{Datasets}

\begin{figure}[!ht]
  \centering
  \includegraphics[width=\linewidth]{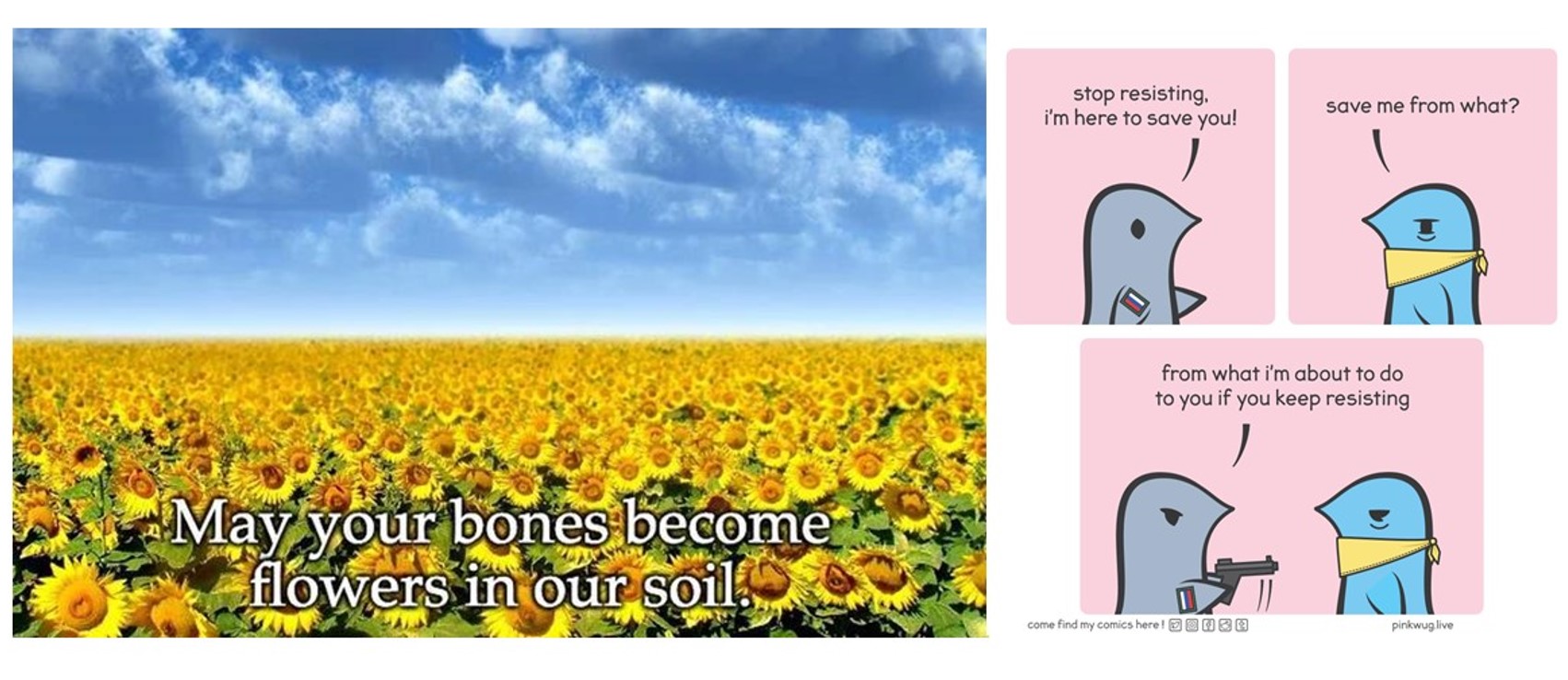}
  \caption{Training data example (Left: NO-HATE, Right: HATE)}
  \label{fig:A}
\end{figure}

In sub-task A, the training dataset provided by the organizers contains 3,600 images. Additionally, a development set and a testing set were provided by the organizers each including 443 instances. Instances in the sub-task A dataset \cite{bhandari2023crisishatemm} are annotated using two labels: NO-HATE (labeled as 0) and HATE (labeled as 1). We present an example of the training data of sub-task A in Figure \ref{fig:A}.

\noindent The label distribution, presented in Table \ref{table: label wise data percentage of ST1}, is skewed in the dataset, with a slightly higher percentage of instances labeled as HATE in the training, testing, and evaluation sets.

\begin{table}[h]

\centering
\begin{tabular}{lcccc}
\hline
\multicolumn{4}{c}{\textbf{sub-task A}} \\
\hline
\textbf{Label} & \textbf{Train} & \textbf{Eval} & \textbf{Test}\\
\hline
\textsc{hate} & 53.95 & 54.85 & 54.85\\
\textsc{no-hate} & 46.05 & 45.15 & 45.15\\
\hline
\end{tabular}
\caption{Distribution of labels in the training, evaluation, and test sets of the sub-task A dataset in terms of percentage.}
\label{table: label wise data percentage of ST1}
\end{table}

\begin{figure}[!ht]
  \centering
  \includegraphics[width=\linewidth]{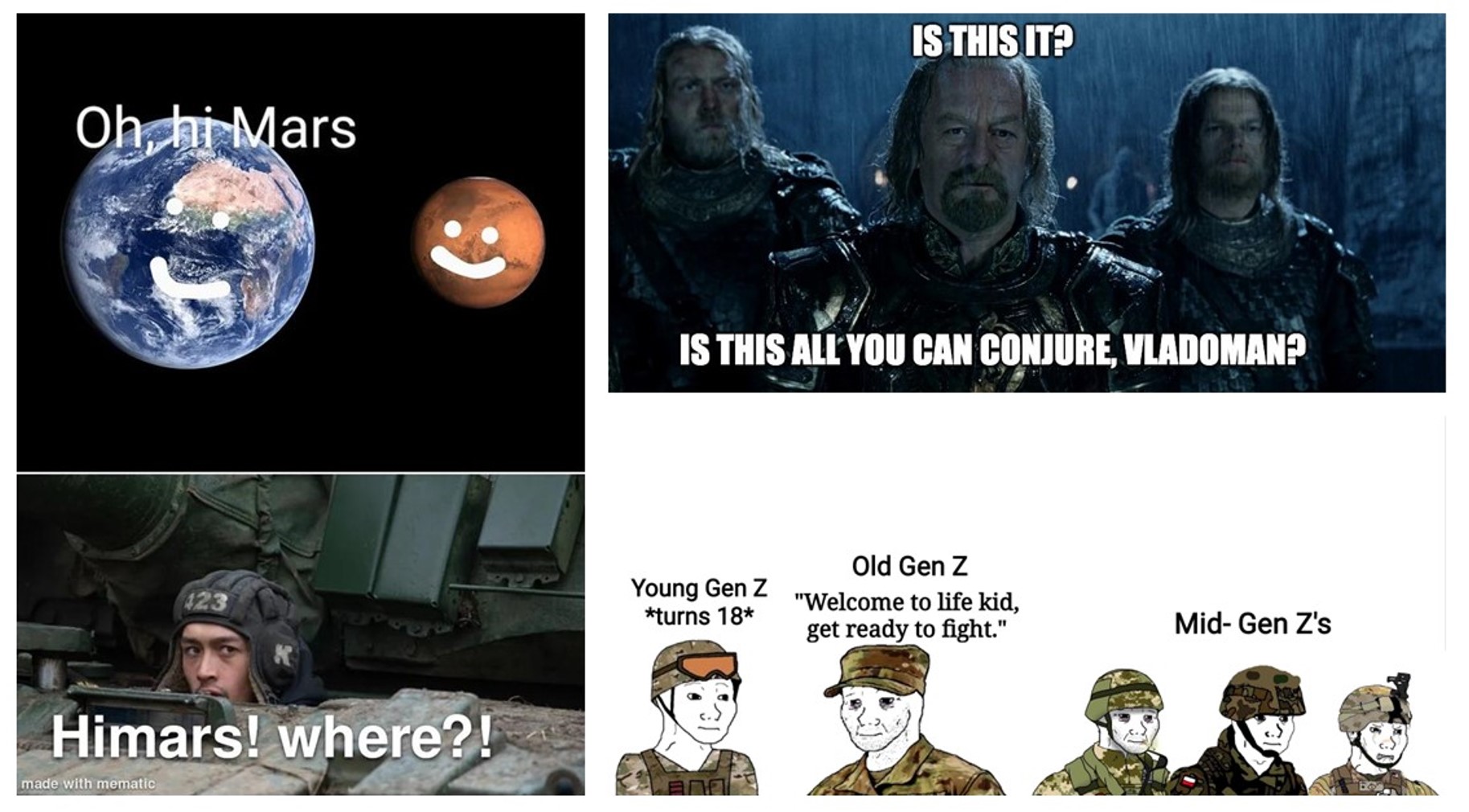}
  \caption{Training data example (Left: Organization, Top-right: Individual, Bottom-right: Community)}
  \label{fig:B}
\end{figure}

\noindent In sub-task B, the training, evaluation, and test sets include 1,942, 244, and 242 images respectively. Instances in the sub-task B dataset \cite{thapa-etal-2022-multi} are labeled into three categories: Individual (labeled as 0), Community (labeled as 1), and Organization (labeled as 2). Examples of training data for sub-task B are shown in Figure \ref{fig:B}.

There is an imbalance among the three labels and the distribution is shown in Table \ref{table:label wise data percentage of ST2}. The class INDIVIDUAL is the most prevalent. The imbalance can impact the model's ability to generalize across different classes, potentially leading to biased results. Addressing this imbalance through techniques like data augmentation or re-balancing strategies may be crucial for developing robust models that perform well across all label categories. 

\begin{table}[h]

\centering
\begin{tabular}{lccc}
\hline
\multicolumn{4}{c}{\textbf{sub-task B}} \\
\hline
\textbf{Label} & \textbf{Train} & \textbf{Eval} & \textbf{Test}\\
\hline
\textsc{individual} & 42.38 & 41.80 & 42.15\\
\textsc{community} & 17.25 & 16.40 & 17.35\\
\textsc{organization} & 40.37 & 41.80 & 40.50\\
\hline
\end{tabular}
\caption{label wise data percentage of sub-task B}
\label{table:label wise data percentage of ST2}
\end{table}

\noindent We have used Google Vision API\footnote{\url{https://cloud.google.com/vision/}} to retrieve text from the images of all the phases of both the sub-tasks. Although the OCR can detect text in a variety of languages, the accuracy may change depending on the language. It's possible that some languages are more accurate and supported than others. The input image quality has an impact on OCR accuracy. 
%Images with a high degree of compression or poor quality may not be as accurate. Accurate text extraction may be difficult in documents with extensive formatting, several columns, or complex layouts. 
In certain situations, the original formatting may not be preserved by the API.

\section{Experiments}

In sub-task A, we use BERTweet-large \cite{berttweetner1} \cite{berttweetner2}, BERT-base \cite{bertbaseuncase}, and XLM-R \cite{xlmr} models. Notably, XLM-R shows the best F1 score. We also use GPT-3.5\footnote{\url{https://platform.openai.com/docs/models/gpt-3-5-turbo}} zero-shot and few-shot prompting with test F1 score 0.73, 0.77. For sub-task B, we also start with BERTweet-large, BERT-base, and XLM-R using the same learning rate and epochs as in sub-task A. Later, we apply a weighted ensemble approach to these models, resulting in the 0.65 F1 score for the task. To tackle class imbalance in sub-task B, we employed back translation, converting the training data through Xosha to Twi to English and Lao to Pashto to Yoruba to English. This significantly improves overall model performance from 0.65 to 0.67.

We follow the approach of back translation of used in \cite{raihan-etal-2023-nlpbdpatriots}. For this, we select languages that demonstrate limited or no cultural overlap with the original language featured in the dataset. Xosha, Twi, Lao, Pashto, and Yoruba are languages that are very diverse culturally and geographically. This diversity underscores the significance of considering a wide range of cultural and geographical influences when working with these languages.
By intentionally selecting these languages without cultural overlap, we introduce a purposeful aspect of diversity, mitigating potential biases, and enhancing the dataset with a broader spectrum of linguistic expressions. Moreover, the Ensemble method with majority voting is also proven helpful in this type of case where a single model may not label the data correctly due to class imbalance \cite{goswami-etal-2023-nlpbdpatriots}. For instance, when two out of three models predict a sentence as a hate event, the sentence is subsequently labeled as a hate event through the application of majority voting. We also use GPT-3.5 zero-shot and few-shot prompting with test F1 scores of 0.53, and 0.57. The prompt provided to GPT3.5 is available in Figure \ref{fig:prompt1}.

\begin{figure}[h]
\centering
\scalebox{.92}{
\begin{tikzpicture}[node distance=1cm]
    % Styles for nodes
    \tikzstyle{block} = [rectangle, draw, fill=green!20, text width=\linewidth, text centered, rounded corners, minimum height=4em]
    \tikzstyle{operation} = [text centered, minimum height=1em]
    % Nodes
    \node [block] (rect1) {\textbf{Role:}{ You are a helpful AI assistant. You are given the task of \textit{<sub-task\_name>}. }};
    \node [operation, below of=rect1] (plus1) {};
    \node [block, below of=plus1] (rect2) {\textbf{Definition:}{ \textit{<sub-task\_definition>}. You will be given a text to label either \textit{<label1>} or \textit{<label2>} or \textit{<label3>}. }};
    \node [operation, below of=rect2] (plus2) {};
    \node [block, below of=plus2] (rect3) {\textbf{Task:}{ Generate the label for this \textbf{text} in the following format: \textit{<label> Your\_Predicted\_Label <$\backslash$label>}. Thanks.}};
\end{tikzpicture}
}
\caption{Sample GPT-3.5 prompt.}
\label{fig:prompt1}
\end{figure}

\noindent We also utilize GPT-3.5 through the OpenAI API for two primary sub-tasks: Hate Speech Detection (sub-task A) and Hate Speech Target Detection (sub-task B). We fine-tune GPT-3.5 using specifically curated training and evaluation datasets, conducting the process over four epochs. It is worth noting that, no other hyper-parameter can be set other than epochs while fine-tuning GPT3.5 through the API. Notably, the OpenAI API does not provide conventional metrics such as training loss, validation loss, precision, or recall. Upon completion of the fine-tuning, the API assigns a unique ID to our model. We use this ID to process the test dataset for both sub-tasks. For labeling and predictions, the API returns results based on the test dataset. In sub-task A, which focuses on detecting hate speech, our model achieves an F1 score of 0.82, indicating a high level of accuracy. Conversely, in sub-task B, where the objective is to identify the targets of hate speech, the model attains a lower F1 score of 0.63, reflecting the inherent challenges in this particular aspect of hate speech analysis.

Hyperparameters of all the models used excluding GPT3.5 in the experiments are available in Figure \ref{tab:training_config}.

\begin{table}[ht]
\centering
\begin{tabular}{lc}
\hline
\textbf{Parameter} & \textbf{Value} \\
\hline
Learning Rate & \(1e-5\) \\
Train Batch Size & 8 \\
Test Batch Size & 8 \\
Epochs & 5 \\
\hline
\end{tabular}
\caption{Training Configuration Parameters}
\label{tab:training_config}
\end{table}

\section{Results}

The detailed experimental results of the models in sub-task A and sub-task B are available in Tables \ref{table:sub-task1_results}, %\ref{table:sub-task1_progress}, 
and \ref{table:sub-task2_results}, 
%\ref{table:sub-task2_progress} 
respectively. In sub-task A, we evaluate a BERT-base, BERTweet-large, and XLM-R model. XLM-R delivers the best performance with a 0.83 F1-score. In sub-task B, our ensemble approach was provides the best F1-score of 0.67.

\begin{table}[h]
\centering
\begin{tabular}{lcc}
\hline
\textbf{Model} & \textbf{Eval F1} & \textbf{Test F1} \\
\hline
\textsc{GPT3.5 (Zero Shot)} & -- & 0.73\\
\textsc{GPT3.5 (Few Shot)} & -- & 0.77\\
\textsc{GPT3.5 (FineTuned)} & 0.86 & 0.82\\
\hline
\textsc{BERT-base} & 0.81 &  0.75\\
\textsc{BERTweet-large} & 0.89 & 0.81\\
\textsc{XLM-R} & 0.95 & \textbf{0.83}\\
\hline
\end{tabular}
\caption{Results of sub-task A.}
\label{table:sub-task1_results}
\end{table}

\newcommand{\progressbar}[2]{%
  \begin{tikzpicture}
    \fill[blue!30!white] (0,0) rectangle (#1\linewidth,0.5);
    \draw (0,0) rectangle (\linewidth,0.5);
    \node at (0.5\linewidth,0.2) {#2\%};
  \end{tikzpicture}%
}

% \begin{table}[h]
% \centering
% \begin{tabular}{c}

%  \progressbar{0.73}{\textsc{GPT3.5 (Zero Shot)} 73} \\
%  \progressbar{0.77}{\textsc{GPT3.5 (Few Shot)} 77} \\
%  \progressbar{0.82}{\textsc{GPT3.5 (FineTuned)} 82} \\
%  \progressbar{0.75}{\textsc{BERT-base} 75} \\
%  \progressbar{0.81}{\textsc{BERTweet-large} 81} \\
%  \progressbar{0.83}{\textsc{XLM-R} 83} \\

% \end{tabular}
% \caption{Test F1 Performance Bar of sub-task A.}
% \label{table:sub-task1_progress}
% \end{table}

\begin{table}[h]
\centering
\resizebox{\linewidth}{!}{%
\begin{tabular}{lcc}
\hline
\textbf{Model} & \textbf{Eval F1} & \textbf{Test F1} \\
\hline
\textsc{GPT3.5 (Zero Shot)} & -- & 0.53\\
\textsc{GPT3.5 (Few Shot)} & -- & 0.57\\
\textsc{GPT3.5 (FineTuned)} & 0.65 & 0.63\\
\hline
\textsc{BERT-base} & 0.61 &  0.60\\
\textsc{XLM-R} & 0.63 & 0.61\\
\textsc{BERTweet-large} & 0.68 & 0.64\\
\hline
\textsc{Ensemble} & 0.69 & 0.65 \\
\hline
\textsc{BERT-base (Aug.)} & 0.63 &  0.61\\
\textsc{XLM-R (Aug.)} & 0.65 & 0.64\\
\textsc{BERTweet-large (Aug.)} & 0.70 & 0.66\\
\hline
\textsc{Ensemble (Aug.)} & 0.71 & \textbf{0.67} \\
\hline
\end{tabular}
}
\caption{Results of sub-task B (before and after data augmentation).}
\label{table:sub-task2_results}
\end{table}

% \begin{table}[h]
% \centering
% \begin{tabular}{c}

% \progressbar{0.53}{\textsc{GPT3.5 (Zero Shot)} 53} \\
%  \progressbar{0.57}{\textsc{GPT3.5 (Few Shot)} 57} \\
%  \progressbar{0.63}{\textsc{GPT3.5 (FineTuned)} 63} \\
%  \progressbar{0.61}{\textsc{BERT-base} 61} \\
%  \progressbar{0.64}{\textsc{XLM-R} 64} \\
%  \progressbar{0.66}{\textsc{BERTweet-large} 66} \\
%  \progressbar{0.67}{\textsc{Ensemble} 67} \\

% \end{tabular}
% \caption{Test F1 Performance Bar of sub-task B.}
% \label{table:sub-task2_progress}
% \end{table}

\section{Error Analysis}

\begin{figure}[!ht]
  \centering
  \includegraphics[width=0.99\linewidth]{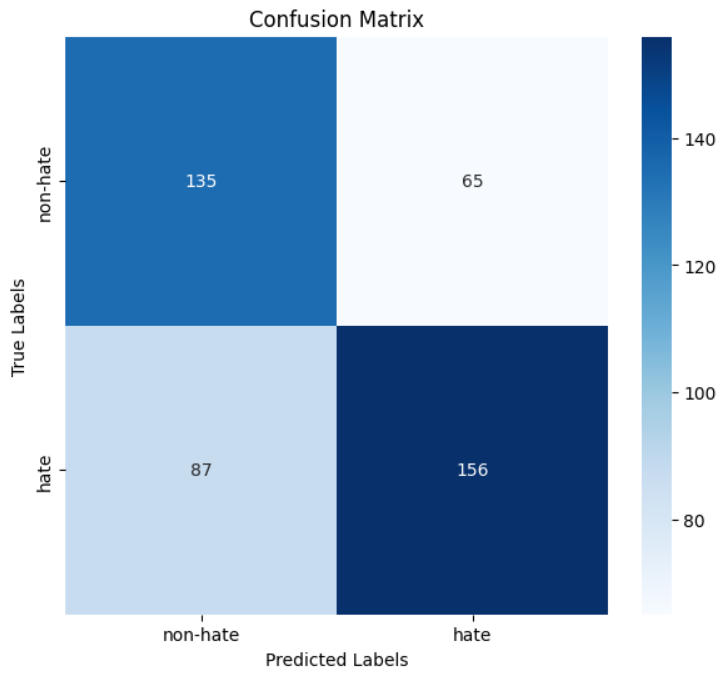}
  \caption{Confusion matrix of sub-task A evaluation set.}
  \label{fig:confusion matrix1}
\end{figure}

In sub-task A, our aim is to detect non-hate (labeled as 0) and hate (labeled as 1) speeches. Therefore, the task of our model is to categorize text into two categories: non-hate or hate. The confusion matrix, presented in Figure \ref{fig:confusion matrix1}, illustrates both the true labels and predicted labels, indicating that our model excels in recognizing hate speech than the non-hate ones. The observed bias towards recognizing hate speech in the model may stem from the prevalence of HATE-labeled texts in both training and evaluation datasets. As both the training and evaluation datasets are used to train the model, the model may develop a bias, impacting its accuracy when dealing with non-hate speeches. 

\begin{figure}[!ht]
  \centering
  \includegraphics[width=0.97\linewidth]{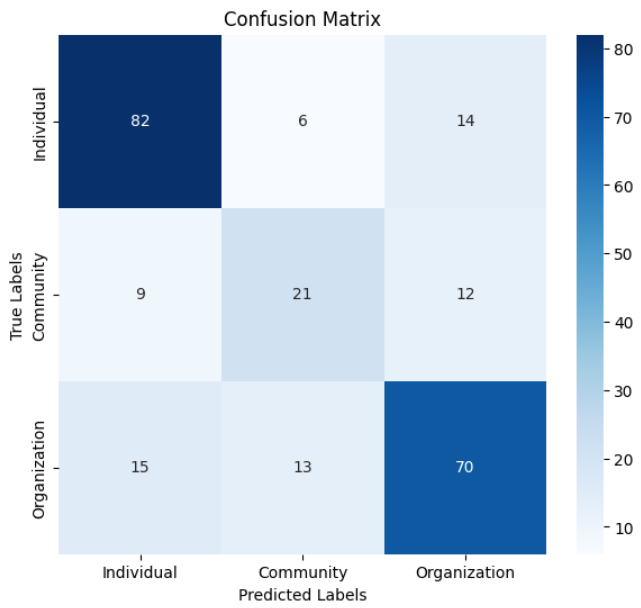}
  \caption{Confusion matrix of sub-task B evaluation set.}
  \label{fig:confusion matrix2}
\end{figure}

\noindent In sub-task B, our ensemble model is assigned the challenge of categorizing targets from text-embedded images into three labels: individual (labeled as 0), community (labeled as 1), and organization (labeled as 2). Analysis of the Confusion Matrix shown in Figure \ref{fig:confusion matrix2}, indicates that our model shows difficulties in identifying community categories, compared to labeling organizations and individuals. However, the model excels in accurately categorizing individuals. This underscores the significance of having a balanced dataset. The observed challenges in the model's performance, particularly in identifying the community category, can be attributed to an imbalance in the training and evaluation datasets. 

According to our initial analysis, there are some challenges that can affect our results. Firstly, there is an imbalance in label distribution within our dataset, where certain data classes contain more Instances than others. This makes it difficult for the model to learn properties of classes that contain fewer examples. %This might make our results a bit biased. 
Secondly, we observed that some labels in the dataset are correctly attributed. This is the case of many offensive and hate speech datasets due to the intrinsic subjectivity of the task, as noted by \citet{weerasooriya2023vicarious}. Incorrect labels can confuse our model, making it harder for it to learn properly and leading to mistakes in the evaluation state. It may also explain why GPT3.5 underperformed, even after finetuning. Also, as this is primarily a text classification task - models like XLM-R do better than GPT3.5. 

Finally, another limitation lies in the impact of external factors on the reliability of our Multimodal Hate Event Detection Model over time. The dynamic nature of online discourse and political shifts may affect its efficacy. Even though our models achieve good results, recognizing and dealing with these challenges is important when developing high-performing models that work well in the ever-changing world of online conversations and political events.

\section{Conclusion and Future Work}

%The employed methodologies marked substantial progress in the field of Multimodal Hate Event Detection. We addressed the difficulties associated with handling multimodal content in our study. The XLM-R model performed exceptionally well in sub-task A, securing the third position with a remarkable Test F1 score of 0.83. In the same way, for sub-task B, our group method, which combined BERT base, BERTweet large, and XLM-R, likewise achieved the third place with test F1 score of 0.67.

This paper evaluated various approaches to Multimodal Hate Event Detection. We tested multiple models such as GPT, XLM-R, and BERT on sub-task a and sub-task b of the competition and we addressed the difficulties associated with handling multimodal content. Our XLM-R model performed well in subtask A ranking third, achieving an F1 score of 0.83. In the same way, for subtask B, our ensemble method, which combined BERT base, BERTweet large, and XLM-R, also ranked third, achieving an F1 score of 0.67.

Despite encountering label distribution imbalances in the training and evaluation sets, our approaches successfully navigated these challenges. Future studies will focus on exploring potential biases in our models and further refining strategies for handling class imbalance as in \citet{akhbardeh2021handling}. Moreover, as online communication continues to increase multimodality, developing robust hate speech detection systems requires fusing information from different modalities. Future work should focus on faceted annotation schemes and semi-supervised approaches to improve generalization. Evaluating model biases, and exploring the impacts of label imbalance are also important areas needing attention. We hope our experiments provide a valuable starting point for further research towards safer online spaces.

\section*{Acknowledgment}

% We would like to thank the shared task organizers for providing us with the dataset used in our study. 

We extend our sincere appreciation to the organizers of the shared task for graciously providing us with this intriguing and valuable dataset. Our heartfelt thanks go to the anonymous reviewers whose insightful feedback has proven invaluable in enhancing the quality of our work. We are truly grateful for the opportunity to engage with such a thought-provoking dataset, and we acknowledge the indispensable role played by the organizers and reviewers in shaping the success of our endeavors.

%Their invaluable cooperation greatly enhanced our study and allowed for a thorough investigation of multimodal hate event identification. The shared task organizers enabled smooth collaboration, which significantly enhanced the scope and depth of our investigation.

%As we advance, we'll be concentrating on improving models to better manage label imbalances and investigating novel approaches for more thorough hate speech detection. Ethics will always come first, and we'll keep working to make sure our models support online discourse in a way that complies with the does not violate any social or virtual ethical aspect.

% \section*{Limitations}

% There are some challenges that can affect our results. Firstly, there exists an imbalance in label distribution within our dataset, where certain data types are more abundant than others. This might make our results a bit biased. Second, the labels (the categories we assign to the data) in our dataset are sometimes wrong or don't match each other. This can confuse our model, making it harder for it to learn properly and leading to mistakes in the results.  Another limitation lies in the impact of external factors on the reliability of our Multimodal Hate Event Detection model over time. The dynamic nature of online discourse and political shifts may affect its efficacy. Even though we've made progress, recognizing and dealing with these challenges is important for our model to work well in the ever-changing world of online conversations and politics.

\section*{Ethics Statement}

This study adheres to the \href{https://www.aclweb.org/portal/content/acl-code-ethics}{ACL Ethics Policy} and seeks to make a contribution to the realm of online safety. The dataset is supplied to us by the organizers and has undergone anonymization to secure the privacy of the users. The technology in question possesses the potential to serve as a beneficial instrument for the moderation of online content, thereby facilitating the creation of safer digital environments. However, it is imperative to exercise caution to prevent its potential misuse for purposes such as monitoring or censorship.

% Entries for the entire Anthology, followed by custom entries
\bibliography{customshort}

\begin{thebibliography}{36}
\expandafter\ifx\csname natexlab\endcsname\relax\def\natexlab#1{#1}\fi

\bibitem[{Akhbardeh et~al.(2021)Akhbardeh, Alm, Zampieri, and Desell}]{akhbardeh2021handling}
Farhad Akhbardeh, Cecilia~Ovesdotter Alm, Marcos Zampieri, and Travis Desell. 2021.
\newblock Handling extreme class imbalance in technical logbook datasets.
\newblock In \emph{Proceedings of ACL}.

\bibitem[{Arag{\'o}n et~al.(2019)Arag{\'o}n, Carmona, Montes-y G{\'o}mez, Escalante, Pineda, and Moctezuma}]{aragon2019overview}
Mario~Ezra Arag{\'o}n, Miguel Angel~Alvarez Carmona, Manuel Montes-y G{\'o}mez, Hugo~Jair Escalante, Luis~Villasenor Pineda, and Daniela Moctezuma. 2019.
\newblock Overview of mex-a3t at iberlef 2019: Authorship and aggressiveness analysis in mexican spanish tweets.
\newblock In \emph{Proceedings of IberLEF}.

\bibitem[{Bhandari et~al.(2023)Bhandari, Shah, Thapa, Naseem, and Nasim}]{bhandari2023crisishatemm}
Aashish Bhandari, Siddhant~B Shah, Surendrabikram Thapa, Usman Naseem, and Mehwish Nasim. 2023.
\newblock Crisishatemm: Multimodal analysis of directed and undirected hate speech in text-embedded images from russia-ukraine conflict.
\newblock In \emph{Proceedings of IEEE CVF}.

\bibitem[{Boishakhi et~al.(2021)Boishakhi, Shill, and Alam}]{Boishakhi_2021}
Fariha~Tahosin Boishakhi, Ponkoj~Chandra Shill, and Md. Golam~Rabiul Alam. 2021.
\newblock Multi-modal hate speech detection using machine learning.
\newblock In \emph{Proceedings of IEEE Big Data}.

\bibitem[{Chiril et~al.(2019)Chiril, Benamara~Zitoune, Moriceau, Coulomb-Gully, and Kumar}]{chiril-etal-2019-multilingual}
Patricia Chiril, Farah Benamara~Zitoune, V{\'e}ronique Moriceau, Marl{\`e}ne Coulomb-Gully, and Abhishek Kumar. 2019.
\newblock Multilingual and multitarget hate speech detection in tweets.
\newblock In \emph{Proceedings of TALN}.

\bibitem[{{\c{C}}{\"o}ltekin(2020)}]{coltekin-2020}
{\c{C}}a{\u{g}}r{\i} {\c{C}}{\"o}ltekin. 2020.
\newblock A corpus of {T}urkish offensive language on social media.
\newblock In \emph{Proceedings of LREC}.

\bibitem[{Conneau et~al.(2020)Conneau, Khandelwal, Goyal, Chaudhary, Wenzek, Guzm{\'a}n, Grave, Ott, Zettlemoyer, and Stoyanov}]{xlmr}
Alexis Conneau, Kartikay Khandelwal, Naman Goyal, Vishrav Chaudhary, Guillaume Wenzek, Francisco Guzm{\'a}n, Edouard Grave, Myle Ott, Luke Zettlemoyer, and Veselin Stoyanov. 2020.
\newblock Unsupervised cross-lingual representation learning at scale.
\newblock In \emph{Proceedings of ACL}.

\bibitem[{Davidson et~al.(2017)Davidson, Warmsley, Macy, and Weber}]{davidson2017}
Thomas Davidson, Dana Warmsley, Michael~W. Macy, and Ingmar Weber. 2017.
\newblock Automated hate speech detection and the problem of offensive language.
\newblock In \emph{Proceedings of ICWSM}.

\bibitem[{Devlin et~al.(2019)Devlin, Chang, Lee, and Toutanova}]{bertbaseuncase}
Jacob Devlin, Ming-Wei Chang, Kenton Lee, and Kristina Toutanova. 2019.
\newblock {BERT}: Pre-training of deep bidirectional transformers for language understanding.
\newblock In \emph{Proceedings NAACL}.

\bibitem[{Fortuna and Nunes(2018)}]{fortuna2018survey}
Paula Fortuna and S{\'e}rgio Nunes. 2018.
\newblock A survey on automatic detection of hate speech in text.
\newblock \emph{ACM Computing Surveys (CSUR)}, 51(4):1--30.

\bibitem[{Gaikwad et~al.(2021)Gaikwad, Ranasinghe, Zampieri, and Homan}]{gaikwad-marathi}
Saurabh~Sampatrao Gaikwad, Tharindu Ranasinghe, Marcos Zampieri, and Christopher Homan. 2021.
\newblock Cross-lingual offensive language identification for low resource languages: The case of {M}arathi.
\newblock In \emph{Proceedings of RANLP}.

\bibitem[{Goswami et~al.(2023)Goswami, Raihan, Puspo, and Zampieri}]{goswami-etal-2023-nlpbdpatriots}
Dhiman Goswami, Md~Nishat Raihan, Sadiya Sayara~Chowdhury Puspo, and Marcos Zampieri. 2023.
\newblock nlp{BD}patriots at {BLP}-2023 task 2: A transfer learning approach towards {B}angla sentiment analysis.
\newblock In \emph{Proceedings of BLP}.

\bibitem[{Grimminger and Klinger(2021)}]{grimminger2021hate}
Lara Grimminger and Roman Klinger. 2021.
\newblock Hate towards the political opponent: A twitter corpus study of the 2020 us elections on the basis of offensive speech and stance detection.
\newblock In \emph{Proceedings of WASSA}.

\bibitem[{Hermida and Santos(2023)}]{hermida2023detecting}
Paulo Cezar de~Q Hermida and Eulanda M~dos Santos. 2023.
\newblock Detecting hate speech in memes: a review.
\newblock \emph{Artificial Intelligence Review}, pages 1--19.

\bibitem[{Ji et~al.(2023)Ji, Ren, and Naseem}]{ji2023identifying}
Junhui Ji, Wei Ren, and Usman Naseem. 2023.
\newblock Identifying creative harmful memes via prompt based approach.
\newblock In \emph{Proceedings of WWW}.

\bibitem[{Karim et~al.(2022)Karim, Dey, Islam, Shajalal, and Chakravarthi}]{karim2022multimodal}
Md~Rezaul Karim, Sumon~Kanti Dey, Tanhim Islam, Md~Shajalal, and Bharathi~Raja Chakravarthi. 2022.
\newblock Multimodal hate speech detection from bengali memes and texts.
\newblock In \emph{Proceedings of SPELL}.

\bibitem[{Mathew et~al.(2021)Mathew, Saha, Yimam, Biemann, Goyal, and Mukherjee}]{mathew2020hatexplain}
Binny Mathew, Punyajoy Saha, Seid~Muhie Yimam, Chris Biemann, Pawan Goyal, and Animesh Mukherjee. 2021.
\newblock Hatexplain: A benchmark dataset for explainable hate speech detection.
\newblock \emph{Proceedings of AAAI}.

\bibitem[{Modha et~al.(2021)Modha, Mandl, Shahi, Madhu, Satapara, Ranasinghe, and Zampieri}]{modha2021overview}
Sandip Modha, Thomas Mandl, Gautam~Kishore Shahi, Hiren Madhu, Shrey Satapara, Tharindu Ranasinghe, and Marcos Zampieri. 2021.
\newblock Overview of the hasoc subtrack at fire 2021: Hate speech and offensive content identification in english and indo-aryan languages and conversational hate speech.
\newblock In \emph{Proceedings of FIRE}.

\bibitem[{Pavlopoulos et~al.(2021)Pavlopoulos, Sorensen, Laugier, and Androutsopoulos}]{pavlopoulos-semeval}
John Pavlopoulos, Jeffrey Sorensen, L{\'e}o Laugier, and Ion Androutsopoulos. 2021.
\newblock {S}em{E}val-2021 task 5: Toxic spans detection.
\newblock In \emph{Proceedings of SemEval}.

\bibitem[{Perifanos and Goutsos(2021)}]{perifanos2021multimodal}
Konstantinos Perifanos and Dionysis Goutsos. 2021.
\newblock Multimodal hate speech detection in greek social media.
\newblock \emph{Multimodal Technologies and Interaction}, 5(7):34.

\bibitem[{Pitenis et~al.(2020)Pitenis, Zampieri, and Ranasinghe}]{pitenis-etal-2020-offensive}
Zesis Pitenis, Marcos Zampieri, and Tharindu Ranasinghe. 2020.
\newblock Offensive language identification in {G}reek.
\newblock In \emph{Proceedings of LREC}.

\bibitem[{Raihan et~al.(2023{\natexlab{a}})Raihan, Goswami, Puspo, and Zampieri}]{raihan-etal-2023-nlpbdpatriots}
Md~Nishat Raihan, Dhiman Goswami, Sadiya Sayara~Chowdhury Puspo, and Marcos Zampieri. 2023{\natexlab{a}}.
\newblock nlp{BD}patriots at {BLP}-2023 task 1: Two-step classification for violence inciting text detection in {B}angla - leveraging back-translation and multilinguality.
\newblock In \emph{Proceedings of BLP}.

\bibitem[{Raihan et~al.(2023{\natexlab{b}})Raihan, Tanmoy, Islam, North, Ranasinghe, Anastasopoulos, and Zampieri}]{raihan2023offensive}
Md~Nishat Raihan, Umma Tanmoy, Anika~Binte Islam, Kai North, Tharindu Ranasinghe, Antonios Anastasopoulos, and Marcos Zampieri. 2023{\natexlab{b}}.
\newblock Offensive language identification in transliterated and code-mixed bangla.
\newblock In \emph{Proceedings of BLP}.

\bibitem[{Rajput et~al.(2022)Rajput, Kapoor, Rai, and Kaur}]{rajput2022hate}
Kshitij Rajput, Raghav Kapoor, Kaushal Rai, and Preeti Kaur. 2022.
\newblock Hate me not: detecting hate inducing memes in code switched languages.
\newblock In \emph{Proceedings of AMCIS}.

\bibitem[{Ranasinghe and Zampieri(2021)}]{ranasinghemudes}
Tharindu Ranasinghe and Marcos Zampieri. 2021.
\newblock {MUDES: Multilingual Detection of Offensive Spans}.
\newblock In \emph{Proceedings of NAACL}.

\bibitem[{Schmidt and Wiegand(2017)}]{schmidt2017survey}
Anna Schmidt and Michael Wiegand. 2017.
\newblock {A Survey on Hate Speech Detection Using Natural Language Processing}.
\newblock In \emph{Proceedings of SocialNLP}.

\bibitem[{Thapa et~al.(2023)Thapa, Jafri, H{\"u}rriyeto{\u{g}}lu, Vargas, Lee, and Naseem}]{thapa-etal-2023-multimodal}
Surendrabikram Thapa, Farhan Jafri, Ali H{\"u}rriyeto{\u{g}}lu, Francielle Vargas, Roy Ka-Wei Lee, and Usman Naseem. 2023.
\newblock Multimodal hate speech event detection - shared task 4, {CASE} 2023.
\newblock In \emph{Proceedings of CASE}.

\bibitem[{Thapa et~al.(2024)Thapa, Rauniyar, Jafri, Veeramani, Jain, Jain, Vargas, H{\"u}rriyeto{\u{g}}lu, and Naseem}]{thapa2024multimodal}
Surendrabikram Thapa, Kritesh Rauniyar, Farhan~Ahmad Jafri, Hariram Veeramani, Raghav Jain, Sandesh Jain, Francielle Vargas, Ali H{\"u}rriyeto{\u{g}}lu, and Usman Naseem. 2024.
\newblock Extended multimodal hate speech event detection during russia-ukraine crisis - shared task at case 2024.
\newblock In \emph{Proceedings of CASE}.

\bibitem[{Thapa et~al.(2022)Thapa, Shah, Jafri, Naseem, and Razzak}]{thapa-etal-2022-multi}
Surendrabikram Thapa, Aditya Shah, Farhan Jafri, Usman Naseem, and Imran Razzak. 2022.
\newblock A multi-modal dataset for hate speech detection on social media: Case-study of russia-{U}kraine conflict.
\newblock In \emph{Proceedings of CASE}.

\bibitem[{Ushio and Camacho-Collados(2021)}]{berttweetner1}
Asahi Ushio and Jose Camacho-Collados. 2021.
\newblock {T}-{NER}: An all-round python library for transformer-based named entity recognition.
\newblock In \emph{Proceedings of EACL}.

\bibitem[{Ushio et~al.(2022)Ushio, Neves, Silva, Barbieri, and Camacho-Collados}]{berttweetner2}
Asahi Ushio, Leonardo Neves, Vitor Silva, Francesco. Barbieri, and Jose Camacho-Collados. 2022.
\newblock {N}amed {E}ntity {R}ecognition in {T}witter: {A} {D}ataset and {A}nalysis on {S}hort-{T}erm {T}emporal {S}hifts.
\newblock In \emph{Proceedings of AACL}.

\bibitem[{Weerasooriya et~al.(2023)Weerasooriya, Dutta, Ranasinghe, Zampieri, Homan, and Khudabukhsh}]{weerasooriya2023vicarious}
Tharindu Weerasooriya, Sujan Dutta, Tharindu Ranasinghe, Marcos Zampieri, Christopher Homan, and Ashiqur Khudabukhsh. 2023.
\newblock Vicarious offense and noise audit of offensive speech classifiers: unifying human and machine disagreement on what is offensive.
\newblock In \emph{Proceedings of EMNLP}.

\bibitem[{Yang et~al.(2022)Yang, Zhu, Liu, Han, and Hu}]{yang2022multimodal}
Chuanpeng Yang, Fuqing Zhu, Guihua Liu, Jizhong Han, and Songlin Hu. 2022.
\newblock Multimodal hate speech detection via cross-domain knowledge transfer.
\newblock In \emph{Proceedings of ACM MM}.

\bibitem[{Zampieri et~al.(2019)Zampieri, Malmasi, Nakov, Rosenthal, Farra, and Kumar}]{zampieri2019predicting}
Marcos Zampieri, Shervin Malmasi, Preslav Nakov, Sara Rosenthal, Noura Farra, and Ritesh Kumar. 2019.
\newblock Predicting the type and target of offensive posts in social media.
\newblock In \emph{Proceedings of NAACL}.

\bibitem[{Zampieri et~al.(2023)Zampieri, Morgan, North, Ranasinghe, Simmmons, Khandelwal, Rosenthal, and Nakov}]{zampieri2023target}
Marcos Zampieri, Skye Morgan, Kai North, Tharindu Ranasinghe, Austin Simmmons, Paridhi Khandelwal, Sara Rosenthal, and Preslav Nakov. 2023.
\newblock Target-based offensive language identification.
\newblock In \emph{Proceedings of ACL}.

\bibitem[{Zia et~al.(2022)Zia, Castro, Zubiaga, and Tyson}]{zia2022improving}
Haris~Bin Zia, Ignacio Castro, Arkaitz Zubiaga, and Gareth Tyson. 2022.
\newblock Improving zero-shot cross-lingual hate speech detection with pseudo-label fine-tuning of transformer language models.
\newblock In \emph{Proceedings of ICWSM}.

\end{thebibliography}
\bibliographystyle{acl_natbib}

\newpage
\appendix

\end{document}